\documentclass[11pt,a4paper, hyphens]{article}
\usepackage[hyperref]{eacl2021}
\usepackage{times}
\usepackage{latexsym}
\usepackage{longtable} 
\usepackage{booktabs}
\usepackage{times}  
\usepackage{helvet} 
\usepackage{courier}  
\usepackage{url}  
\usepackage{graphicx} 
\usepackage[bottom]{footmisc}
\usepackage{soul}
\usepackage{natbib}
\usepackage{longtable} 
\usepackage{booktabs}
\usepackage{makecell}
\usepackage{amsmath}
\newcommand{\name}{\texttt{LESA}}
\usepackage{microtype}

\aclfinalcopy 

\title{\name: Linguistic Encapsulation and Semantic Amalgamation Based Generalised Claim Detection from Online Content}

\author{Shreya Gupta$^\dagger{}^*$, Parantak Singh$^\ddagger$\thanks{$^*$ First two authors have equal contributions. The work was done when Parantak was an intern at LCS2 Lab, IIIT-Delhi.} , Megha Sundriyal$^\dagger$, \\ \textbf{Md Shad Akhtar$^\dagger$, Tanmoy Chakraborty$^\dagger$} \\ \\
  $^\dagger$ \textit {IIIT-Delhi, India.}
  $^\ddagger$ \textit{Birla Institute of Technology and Science, Pilani, Goa, India.} \\
  \texttt{\{shreyag, meghas, shad.akhtar, tanmoy\}@iiitd.ac.in}, \\ \texttt{f20170109@goa.bits-pilani.ac.in}
}

\date{}

\usepackage{enumitem,kantlipsum}
\usepackage{amsmath,amsfonts,amssymb}
\usepackage{graphicx,accents}
\usepackage{natbib}
\usepackage{multirow}

\begin{document}
\maketitle
\begin{abstract}
The conceptualization of a claim lies at the core of argument mining. The segregation of claims is complex, owing to the divergence in textual syntax and context across different distributions. Another pressing issue is the unavailability of {\em labeled} unstructured text for experimentation. In this paper, we propose \name, a framework which aims at advancing headfirst into expunging the former issue by assembling a source-independent generalized model that captures syntactic features through part-of-speech and dependency embeddings, as well as contextual features through a fine-tuned language model. We resolve the latter issue by annotating a Twitter dataset which aims at providing a testing ground on a large unstructured dataset.
Experimental results show that \name\ improves upon the state-of-the-art performance across six benchmark claim datasets by an average of 3 claim-F1 points for in-domain experiments and by 2 claim-F1 points for general-domain experiments. On our dataset too, \name\ outperforms existing baselines by 1 claim-F1 point on the in-domain experiments and 2 claim-F1 points on the general-domain experiments. We also release comprehensive data annotation guidelines compiled during the annotation phase (which was missing in the current literature).
\end{abstract}

\section{Introduction} 

The concept of a claim lies at the core of 
the argument mining task. \citet{toulmin2003uses}, in his argumentation theory, described the term `claim' as `{\em an assertion that deserves our attention}’; albeit not very precise, it still serves as an initial insight. In recent years, \citet{govier2013practical}  described a `claim' as `{\em a disputed statement that we try to support with reasons}.' 

The predicament behind the claim detection task exists given the disparity in conceptualization and lack of a proper definition of a claim. The task of claim detection across different domains has garnered tremendous attention so far owing to an uprise in social media consumption and by extension the existence of fake news, online debates, widely-read blogs, etc. As an elementary example, claim detection can be used as a precursor to fact-checking; wherein segregation of claims aids in restricting the corpus that needs a fact-check. A few examples are shown in Table \ref{tab:claim:exm}.

Most of the existing works are built upon two fundamental pillars - semantic encapsulation \citep{daxenberger2017essence, chakrabarty2019imho} and syntactic encapsulation \citep{levy2014context, lippi2015context}. They mainly focus on adapting to texts from similar distributions or topics or both. Secondly, they often exercise against well-structured and laboriously pre-processed formal texts owing to the lack of a labeled corpus consisting of unstructured texts. As a result, claim detection from unstructured raw data still lies under a relatively less explored umbrella.    

\begin{table}[h]
    \centering
    
     \resizebox{0.9\columnwidth}{!}
    {
    \begin{tabular}{p{15em}|c}
         \multicolumn{1}{c|}{\bf Text} & \bf Claim? \\ \hline
        
        \hline
         Alcohol cures corona. & Yes \\ \hline
         Wearing mask can prevent corona. & Yes \\ \hline
         Lord, please protect my family \& the Philippines from the corona virus. & \multirow{2}{*}{No} \\ \hline
         If this corona scare doesn't end soon imma have to intervene & \multirow{2}{*}{No} \\\hline
         
         \hline
    \end{tabular}}
    \caption{A few examples of claim and non-claim.}
    \label{tab:claim:exm}
\end{table}

\noindent \textbf{Motivation:} Claims can be sourced from a variety of sources, e.g., online social media texts, microblogs, Wikipedia articles, etc. It is, however, crucial to pay special attention to claims observed on online social media (OSM) sites \cite{covidreport, who}. Twitter, being a major OSM platform, provides the perfect playground for different ideologies and perspectives. 
Over time, Twitter has emerged as the hub for short, unstructured pieces of text that describe anything from news to personal life. 
Most individuals view and believe things that align with their compass and prior knowledge, {\em aka conformity bias} \cite{whalen2015conformity} -- users tend to make bold claims that usually create a clash between users of varied opinions. At times, these claims incite a negative impact on individuals and society. As an example, a tweet that reads “\textit{alcohol cures corona}” can lead to massive retweeting and consequential unrest, especially in times of a pandemic, when people are more vulnerable to suggestions. In such cases, automated promotion of claims for immediate further checks could prove to be of utmost importance. An automated system is pivotal since OSM data is far too voluminous to allow for manual human checks, even if it was an expert.


At the same time deploying separate systems contingent on the source of a text is inefficient and moves away from the goal of attaining human intelligence in natural language processing tasks. An ideal situation would be a framework that can effectively detect claims in the general setting. 
However, a major bottleneck towards this goal is the unavailability of an annotated dataset from noisy platforms like Twitter. We acknowledge this bottleneck and, in addition to proposing a generalised framework, we develop a qualitative annotated resource and guidelines for claim detection in tweets.      

\noindent{\bf Proposed Method:} There exists several claim detection models; however, the downside is that most of them are trained on structured text from a specific domain. 
Therefore, in this work, we propose \textbf{\name}, a {\bf L}inguistic {\bf E}ncapsulation and {\bf S}emantic {\bf A}malgamation based generalized claim detection model that is capable of accounting for different text distributions, simultaneously. To formalize this, we divide the text, contingent upon their structure, into three broad categories -- noisy text (\textit{tweets}), semi-noisy text (\textit{comments}), and non-noisy text (\textit{news, essays}, etc.). We model each category separately in a joint framework and fuse them together using attention layers. 

Since the task of claim detection has a strong association with the structure of the input, as argued by \citet{lippi2015context}, we leverage two linguistic properties -- part-of-speech (POS) and dependency tree, 
to capture the linguistic variations of each category. Subsequently, we amalgamate these features with BERT \citep{devlin2018bert} for classification. 

We evaluate \name\ on seven different datasets (including our Twitter dataset) 
and observe efficient performance in each case. Moreover, we compare \name's performance against various state-of-the-art systems for all seven datasets in the general and individual settings. The comparative study advocates the superior performance of \name.

\noindent \textbf{Summary of the Contributions:}
We summarize our major contributions below:
\begin{itemize}[leftmargin=*]
    \item {\bf Twitter claim detection dataset and comprehensive annotation guidelines.} To mitigate the unavailability of an annotated dataset for claim detection in Twitter, we develop a large COVID-19 Twitter dataset, the first of its kind, with $\sim10,000$ labeled tweets, following a comprehensive set of claim annotation guidelines.
    \item {\bf \name, a {\em generalized} claim detection system.} We propose a generalized claim detection model, \name, that identifies the presence of claims in {\em any} online text, without prior knowledge of the source and independent of the domain. To the best of our knowledge, this is the first attempt to define a model that handles claim detection from both structured and unstructured data in conjunction.
    \item {\bf Exhaustive evaluation and superior results.} We evaluate \name\ against multiple  state-of-the-art models on six benchmark claim detection datasets and our own Twitter dataset. Comparison suggests \name's superior performance across datasets and the significance of each model component.   
\end{itemize}

    
    
    

\noindent{\bf Reproducibility:} Code and dataset is publicly available at \url{https://github.com/LCS2-IIITD/LESA-EACL-2021}. Appendix comprises of detailed dataset description, annotation guidelines, hyper-parameters, and additional results.

\section{Related Work} 

In the past decade, the task of claim detection has become a popular research area in text processing with an initial pioneering attempt by \citet{rosenthal2012detecting}. They worked on mining claims from 
discussion forums and employed a supervised approach with features based on sentiment and word-grams.
%
 \citet{levy2014context} proposed a context dependent claim detection (CDCD) approach. They described 
 CDC as `\textit{a general, concise statement that directly supports or contests the given topic.'} 
Their approach was evaluated over Wikipedia articles; 
it detected sentences that include CDCs using context-based and context-free features. This was followed by ranking and detecting CDCs using 
logistic regression. 
\citet{lippi2015context} proposed context-independent claim detection (CICD) using linguistic reasoning, and encapsulated structural information to detect claims. They used constituency parsed trees to extract structural information and predicted parts of the sentence holding a claim using SVM. Although their approach achieved promising results, 
they also used a Wikipedia dataset which was highly engineered and domain dependent.  

\citet{daxenberger2017essence} used six disparate datasets and contrasted the performance of several supervised models. They performed two sets of experiments -- in-domain CD (trained and tested on the same dataset) and cross-domain CD (trained on one and tested on another unseen dataset). They learned divergent conceptualisations of claims over cross-domain datasets. 
\citet{levy2017unsupervised} proposed the first unsupervised approach for claim detection. They  hypothesised a ``claim sentence query'' as an ordered triplet: $\langle${that} $\rightarrow$ {MC} $\rightarrow$ {CL}$\rangle$. According to the authors, a claim begins with the word `that' and is followed by the main concept (MC) or topic name which is further followed by words from a pre-defined claim lexicon (CL). 
This approach would not fit well for text stemming from social media platforms owing to a lack of structure and the use of `that' as an offset for claim.

In recent years transformer-based language models have been employed for claim detection. \citet{chakrabarty2019imho} used over 5 million self-labeled Reddit comments that contained the abbreviations IMO (In My Opinion) or IMHO (In My Honest Opinion) 
to fine-tune their model. However, they made no attempt to explicitly encapsulate the structure of a sentence.

Recently, the CLEF-2020 shared task \citep{checkthat:clef:2020} attracted multiple models which are tweaked specifically for claim detection. \citet{williams2020accenture} bagged the first position in the task using a fine-tuned RoBERTa \cite{liu2019roberta} model with mean pooling and dropout. First runner up of the challenge, \citet{nikolov2020team} used logistic regression on various meta-data tweet features and a RoBERTa-based prediction. \citet{cheema2020checksquare}, the second runner up,  incorporated pre-trained BERT embeddings along with POS and dependency tags as features trained using SVM. 


Traditional approaches focused primarily on the syntactic representations of claims 
and textual feature generation, while recent neural methods leverage transformer models.
With \name, we attempt to learn from the past while building for the future -- we propose encapsulating syntactic representations in the form of POS tags and dependency sequences along with the semantics of the input text using transformer-based BERT \cite{devlin2018bert}. Another key observation has been the use of highly structured and domain-engineered datasets for training the existing models in claim detection. In the current age of alarming disinformation, we recognise the augmented need for a system that can detect claims in online text independent of its origin, context or domain. Therefore, in addition to considering texts from different online mediums, we incorporate, for the first time, a self-annotated large Twitter dataset to the relatively structured datasets that exist in this field. 

\begin{figure*}
    \centering
    \includegraphics[width=1\textwidth]{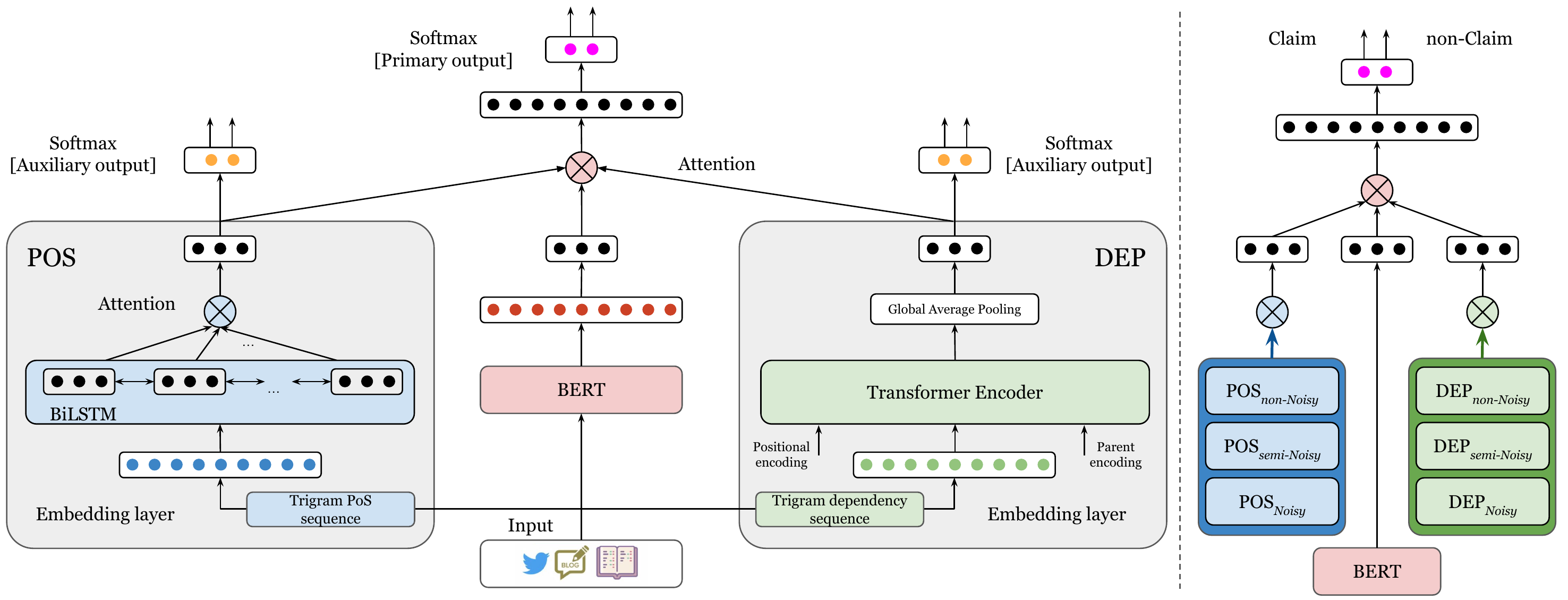}
    \caption{Schematic diagram of our proposed \name\ model. The structure on the right is a high level schematic diagram. Structure on the left shows POS and DEP for one viewpoint.}
    \label{fig:architecture}
\end{figure*}
\section{Proposed Methodology}

Traditionally, the narrative on claim detection is built around either syntactic \cite{levy2017unsupervised, lippi2015context} or semantic \cite{daxenberger2017essence, chakrabarty2019imho} properties of the text. However, given our purview on the integration of both, we propose a combined model, \name\ , that incorporates exclusively linguistic features leveraged from part-of-speech (POS) tags and dependency tree (DEP) as well as semantic features leveraged from transformer-based model, BERT \cite{devlin2018bert}. 

By the virtue of digital media, we generally deal with texts from three kind of environments: (a) a controlled platform where content is pre-reviewed (e.g., news, essays, etc.); (b) a free platform where authors have the freedom to express themselves without any restrictions on the length (e.g., online comments, Wikipedia talk pages); and (c) a free platform with restrictions on the text length (e.g., tweets). The texts in the first category is usually free of any grammatical and typographical mistakes, and thus belong to the non-noisy category. On the other hand, in the third case, texts exhibit a significant amount of noise, in terms of spelling variations, hashtags, emojis, emoticons, abbreviations, etc., to express the desired information within the permissible limit, thus it belongs to the noisy class. The second case is a mixture of the two extreme cases and hence, constitutes the semi-noisy category. We employ three pre-trained models representing noisy, semi-noisy, and non-noisy data for both POS and dependency-based features. The intuition is to leverage the structure-specific linguistic features in a joint framework.

Domain adaptation from a structured environment to an unstructured one is non-trivial and requires specific processing. Therefore, to ensure generalization, we choose to process each input text from three different viewpoints (structure-based segregation), and intelligently select the contributing features among them through an attention mechanism. 
We use it to extract the POS and DEP-based linguistic features. Subsequently, we fuse the linguistic and semantic features using another attention layer before feeding it to a multilayer perceptron (MLP) based classifier. The idea is to amalgamate diverse set of features from different perspectives and leverage them for the final classification. A high-level architectural diagram is depicted in Figure \ref{fig:architecture}. We design parallel pillars for each viewpoint (right side of Figure \ref{fig:architecture}) such that the noisy pillar contains 
pre-trained information from the noisy source and so on. When the common data is passed through the three pillars we hypothesize each pillar’s contribution dependending on the type of input data. For example, if the data source is from a noisy platform, we hypothesize that the noisy pillars will have more significance than the other two viewpoints. We demonstrate this effect in Table \ref{tab:ablation}.

\subsection{Part-of-speech (POS) Module}
The POS module consists of an embedding layer followed by a BiLSTM and an attention layer to extract the syntactic formation of the input text. We pre-train the POS module for each viewpoint, and later fine-tune them while training the integrated model.    

At first, each sequence of tokens $\{x_1, x_2, ..., x_n\}$ is converted to a sequence of corresponding POS tags resulting into the set $\{p_1, p_2, ..., p_n\}$. However, the foremost limitation of this modeling strategy is the limited and small vocabulary size of $19$ owing to a specific number of POS tags. To tackle this, we resort to using $k$-grams of the sequence. The sequence of POS tags (with $k=3$) now becomes $\{(p_0, p_1, p_2), (p_1, p_2, p_3), (p_2, p_3, p_4), ..., $ $(p_{n-2}, p_{n-1}, p_n), (p_{n-1}, p_n, p_{n+1})\}$, where $p_0$ and $p_{n+1}$ are dummy tags. Subsequently, a skip-gram model \cite{mikolov2013efficient} is trained on the POS-transformed corpus of each dataset, which thereby translates to a POS embedding, $E_P$.

\subsection{Dependency Tree (DEP) Module}
Dependency parsing is the function of abstracting the grammatical assembly of a sequence of tokens $\{x_1, x_2, ..., x_n\}$ such that there exists a directed relation (dependency), $d(x_i, x_j)$, between any two tokens $x_i$ and $x_j$, where $x_i$ is the headword and $x_j$ is modified by the headword. We obtain these dependency relations through spaCy\footnote{\url{www.spacy.io}} which uses the clearNLP guidelines. Initially, each sequence is rendered into a combination of the \textit{dependency-tag} arrangement $\{d_1, d_2, ..., d_n\}$ and a \textit{parent-position} arrangement $\{pp_1, pp_2,... pp_n\}$. Here, each $d_j$ represents a dependency tag, where $x_j$ is modified by $x_i$, and $pp_j$ is the index of the modifier (headword) $x_i$. 

We then leverage the transformer encoder \cite{vaswani2017attention}, where traditionally, a position-based signal is added to each token’s embedding to help encode the placement of tokens. In our modified version, the token sequence is the \textit{dependency-tag} sequence $d_e = \{d_1, d_2, ..., d_n\}$, wherein a \textit{parent-position} based signal is additionally added to encode the position of the modifier words. 
\begin{eqnarray}
d'_e = d_e + [(E_{p_1},E_{pp_1}), ..., (E_{p_n},E_{pp_n})]
\end{eqnarray}
where $d'_e \in \mathbf{R}^{d \times n}$ is the modified dependency embedding of a sequence of length $n$, $E_{p_i}$ and $E_{pp_i}$ are the encodings for the \textit{token-position} and the \textit{parent-position} (position of token's modifier), and $(,)$ represent tuple brackets.

This helps us create a flat representation for a dependency graph. The transformer-architecture that we employ comprises of 5 attention heads with an embedding size of 20. Given that there are only a handful of dependency relations, this, still poses the problem of a limited vocabulary size of 37. Having accounted for the parent positions already, we decide to again employ tri-gram sequences $\{(d_0, d_1, d_2), (d_1, d_2, d_3), (d_2, d_3, d_4), ..., $ $(d_{n-2}, d_{n-1}, d_{n}), (d_{n-1}, d_{n}, d_{n+1})\}$ in place of uni-grams.

\section{Datasets}
\label{sec:dataset}

A number of datasets exist for the task of claim detection in online text \cite{peldszus-stede-2015-joint,stab-gurevych-2017-parsing}; however, most of them are formal and structured texts. As we discussed earlier, OSM platforms  are overwhelmed with various claim-ridden posts. Despite the abundance of tweets, literature does not suggest any significant effort for detecting claims in Twitter; Arguably, the prime reason is the lack of a large-scale dataset. Recently, a workshop on claim detection and verification in Twitter was organized under CLEF-2020 \cite{checkthat:clef:2020}. It had two subtasks related to claim identification with separate datasets. The first dataset consists of $1,060$ COVID-19 tweets for claim detection; whereas, the second one comprises of another $1,000$ tweets for claim retrieval. In total, there were $2,069$ annotated tweets of which $1,704$ had claims and $365$ were non-claims. Another recent in-progress dataset on claim detection, which currently has only $305$ claim and $199$ non-claim tweets, was released by \citet{alam2020fighting}. 


Unfortunately, the aforementioned limited instances are insufficient to develop an efficient model. 
Therefore, we attempt to develop a new and relatively larger dataset for claim detection in OSM platforms. We collected $\sim 40,000$ tweets from various sources \cite{carlson_2020,smith_2020,celin_2020,
info:doi/10.2196/19273,QCRI} and manually annotated them. We additionally included claim detection datasets of \citet{alam2020fighting} and CLEF-2020 \cite{checkthat:clef:2020} and re-annotated them in accordance with our guidelines. During the cleaning process, we filtered a majority of tweets due to their irrelevancy and duplicacy. To ensure removal of duplicates, we performed manual checking and exhaustive preprocessing.



\begin{table}[t]
    \centering
    \resizebox{0.3\textwidth}{!}
    {
    \begin{tabular}{l|cc}
    
    & \multicolumn{2}{c}{Our Annotation}\\ \cline{2-3}
CLEF-2020 & Non-claim & Claim \\ \hline

\hline
Non-claim   & 301 & 47 \\
Claim   & 64 & 550 \\ \hline

    \end{tabular}}
    \caption{Confusion matrix highlighting the differences and similarities between \citet{alam2020fighting} and our annotation guidelines for CLEF-2020 claim dataset.}
    \label{tab:reannotation}
\end{table}

\begin{table*}[ht!]
    
    \centering
    \resizebox{0.9\textwidth}{!}{
    \begin{tabular}{l|l|p{38em}}
         \multicolumn{2}{c|}{{\bf Dataset}} & \multicolumn{1}{c}{\bf Text} \\ \hline
        
        \hline
         \multirow{2}{*}{\bf Noisy} & \multirow{2}{*}{TWR} & 
         \underline{@realDonaldTrump} Does ingesting bleach and shining a bright light in the rectal area really cure \underline{\#COVID19}? Have you tried it? Is that what killed Kim Jong Un? \underline{\#TrumpIsALaughingStock \#TrumpIsALoser}
         \\\hline
         \multirow{4}{*}{\bf Semi-noisy} & \multirow{3}{*}{OC} &
         \underline{*smacks blonde wig on Axel*} I think as far as \underline{DiZ} is concerned, he is very smart but also in certain areas very dumb - - witness the fact that he didn't notice his apprentices were going to turn on him, when some of them \underline{(cough Vexen cough)} aren't exactly subtle by nature. \\ \cline{2-3}
         & \multirow{1}{*}{WTP} & \underline{Not to mention one} without any anonymous users TALKING IN CAPITAL LETTERS \underline{!!!!!!!!}
         \\\hline
         \multirow{4}{*}{\bf Non-noisy} & \multirow{1}{*}{MT} & Tax data that are not made available for free should not be acquired by the state. \\ \cline{2-3}
         & \multirow{1}{*}{PE} & I believe that education is the single most important factor in the development of a country. \\ \cline{2-3}
         & \multirow{1}{*}{VG} & When's the last time you slipped on the concept of truth? \\\cline{2-3}
         & \multirow{1}{*}{WD} & The public schools are a bad place to send a kid for a good education anymore. \\ \hline
 
    \end{tabular}}
    \caption{One example from each dataset. Underlined text highlights noisy and semi-noisy phrases.}
    \label{tab:all:datasets:exm}
\end{table*}

\noindent\textbf{Data Annotation:}
To annotate the tweets, we extend and adapt the claim annotation guidelines of \citet{alam2020fighting}. The authors targeted and annotated only a subset of claims, i.e., factually verifiable claims. They did not consider personal opinions, sarcastic comments, implicit claims, or claims existing in a sub-sentence or sub-clause level. Subsequently, we propose our definition of claims and extrapolate the existing guidelines to be more inclusive, nuanced and applicable to a diverse set of claims. Our official definition for claims, adopted from Oxford dictionary\footnote{\tt \url{https://www.lexico.com/definition/claim}}, is to {\em state or assert that something is the case, with or without providing evidence or proof}. 

We present the details of annotation guidelines in \newcite{gupta:lesa:claim:detection:eacl:2021:supplementary}. Following the guidelines, we annotated the collected tweets, and to ensure coherence and conformity, we re-annotated the tweets of \citet{alam2020fighting} and CLEF-2020 \cite{checkthat:clef:2020}. It is intriguing to see the differences and similarities of the two guidelines; therefore, we compile a confusion matrix for CLEF-2020 claim dataset, as presented in Table \ref{tab:reannotation}. Each tweet in our corpus of $9,894$ tweets has been annotated by at least two annotators, with an average Cohen's kappa inter-annotator agreement \cite{cohen1960} score of $0.62$. In case of a disagreement, the third annotator was considered and a majority vote was used for the final label. All annotators were linguists.

\begin{table}[t]
    \centering
    \resizebox{0.4\textwidth}{!}
    {
    \begin{tabular}{c|c|c|c|c|c|c|c|c}
      \multicolumn{2}{c|}{\multirow{2}{*}{\bf Dataset}} & \multicolumn{1}{c|}{\bf Noisy} & \multicolumn{2}{c|}{\bf Semi-noisy} & \multicolumn{4}{c}{\bf Non-noisy} \\ \cline{3-9} 
      & & \textbf{TWR} & \textbf{OC} & \textbf{WTP} & \textbf{MT} & \textbf{PE} & \textbf{VG} & \textbf{WD}\\ \hline 
         
         \hline
         \multirow{2}{*}{Tr} & Cl & 7354 & 623 &1030 & 100 & 1885 & 495 & 190   \\
          & N-cl & 1055 & 7387 & 7174 & 301 & 4499 & 2012 & 3332 \\ \hline
          \multirow{2}{*}{Ts} & Cl & 1296 & 64 & 105 & 12 & 223 & 57 & 14  \\
          & N-cl & 189 & 730 & 759 & 36 & 509 & 221 & 221 \\ \hline
          \multirow{2}{*}{Tot} & Cl & 8650 & 687 & 1,135 & 112 & 2,108 & 552 & 204 \\
          & N-cl & 1244 & 8117 & 7933 & 337 & 5008 & 2233 & 3553 \\\hline 
 
          \hline
    \end{tabular}
    }
     \caption{Statistics of the datasets (Abbreviations: Cl: Claim, N-Cl: Non-claim, Tr: Train set, Ts: Test set; Tot: Total).}
    \label{tab:dataset:stat}
\end{table}
\noindent\textbf{Other Datasets:} \label{other datasets} 
Since we attempt to create a generalized model that is able to detect the presence of a claim in any online text, we accumulate, in addition to the Twitter dataset, six publicly available benchmark datasets:
(i) Online Comments (OC) containing Blog threads of LiveJournal \cite{6061427}, 
(ii) Wiki Talk Pages (WTP) \cite{article},
(iii) German Micro-text (MT) \cite{peldszus-stede-2015-joint},
(iv) Persuasive Student Essay (PE) \cite{stab-gurevych-2017-parsing},
(v) Various Genres (VG) containing  newspaper editorials, parliamentary records and judicial summaries,
and (vi)  Web Discourse (WD) containing blog posts or user comments \cite{habernal-gurevych-2015-exploiting}.
All datasets utilised in this paper contain English texts only. For German Microtexts (MT), we used the publicly available English translated version published by MT’s original authors \cite{peldszus-stede-2015-joint}). The same was utilized by \citet{chakrabarty2019imho}.

The datasets are formed by considering text at the sentence level. For example, in Persuasive Essays (PE) dataset, each essay is broken into sentences and each sentence is individually annotated for a claim. Considering the structure of the input texts in these datasets, we group them into three categories as follows: Noisy (Twitter), Semi-noisy (OC, WTP), Non-noisy (MT, PE, VG, WD). We list one example from each dataset in Table \ref{tab:all:datasets:exm}. 
We also highlight the noisy and semi-noisy phrases in Twitter, and OC and WTP datasets respectively. Moreover, we present detailed statistics of all seven datasets in Table \ref{tab:dataset:stat}.

\section{Experimental Setup}\label{sec:setup}
For all datasets besides twitter, we use the train, validation, and test splits as provided by UKP Lab\footnote{\label{footnote:ukpdataset}\tt \url{https://tinyurl.com/yyckv29p}}.
A mutually exhaustive 70:15:15 split was maintained for Twitter dataset. We compute POS embeddings by learning word2vec skip-gram model \cite{mikolov2013efficient} on the tri-gram\footnote{Choice of $n=3$ is empirical. We report supporting experimental results in \newcite{gupta:lesa:claim:detection:eacl:2021:supplementary}.} POS sequence. For the skip-gram model, we set \textit{context window} $=6$, \textit{embedding dimension} $=20$, and discard the POS sequence with \textit{frequency} $\le2$.
Subsequently, we compute dependency embeddings with \textit{dimension} $=20$ using Transformer \cite{vaswani2017attention} encoder with $5$ \textit{attention heads}. Please note that the choice of using Bi-LSTM, as oppose to Transformers, for extracting the POS features is empirical\footnote{\newcite{gupta:lesa:claim:detection:eacl:2021:supplementary} accompanies the supporting results.}. 

\begin{table*}[t]

\centering
\resizebox{0.9\textwidth}{!}{
\begin{tabular}{l||c|c|c|c|c|c|c|c|c|c|c|c|c|c||c|c}
   \multirow{3}{*}{\bf Models} & \multicolumn{2}{c|}{\bf Noisy} & \multicolumn{4}{c|}{\bf Semi-Noisy} & \multicolumn{8}{c||}{\bf Non-Noisy} & \multicolumn{2}{c}{\multirow{3}{*}{\bf Wt Avg}} \\ 
\cline{2-15}
 & \multicolumn{2}{c|}{Twitter} & \multicolumn{2}{c|}{OC} & \multicolumn{2}{c|}{WTP} & \multicolumn{2}{c|}{MT} & \multicolumn{2}{c|}{PE} & \multicolumn{2}{c|}{VG} & \multicolumn{2}{c||}{WD} & \multicolumn{2}{c}{} \\ \cline{2-15}
 & $m\text{-}F1$ & $c\text{-}F1$ & $m\text{-}F1$ & $c\text{-}F1$& $m\text{-}F1$ & $c\text{-}F1$& $m\text{-}F1$ & $c\text{-}F1$& $m\text{-}F1$ & $c\text{-}F1$& $m\text{-}F1$ & $c\text{-}F1$& $m\text{-}F1$ & $c\text{-}F1$   \\ \hline
 
\hline
BERT & 0.60 &	0.83 &	0.52 &	\bf0.24 &	0.53& \bf0.32&	0.70 &	0.63&	0.69&	0.64&	0.58&	\bf0.43&	0.48&	0.22& 0.58 	& 0.73 \\
\hline

BERT + POS & 0.61 &	0.84 &	\bf 0.53 &	\bf0.24 &	0.54 &	0.31 &	0.75 &	0.69 &	0.72 &	0.64 &	0.59 &	\bf0.43 &	0.51 &	0.24 &	0.60 &	0.74\\

BERT + Dependency & 0.59 &	0.82 &	0.51 &	0.23 &	0.52&	0.30 &	\bf0.79 &	\bf0.73 &	0.69&	0.62&	0.56&	0.41&	0.48&	0.22&	0.57 &	0.72\\ 

POS + Dependency & 0.45 &	0.70 &	0.48 & 0.19 &	 0.47 &	0.25 &	0.57 &	0.46 &	0.50 &	0.45 &	0.56 & 0.41 & 0.44 &	0.17 & 0.48 & 0.61\\

\hline
\name\ (Combined-view) &	0.61 &	0.85&	0.51&	0.23&	0.53&	0.31&	0.77&	0.71&	0.71&	0.64&	0.57&	0.40&	0.48&	0.22&  0.59  & \bf0.75 \\

\hline

\name (768\textit{dim}) & 0.58 &  0.80 & 	0.52 & \bf0.24	 & \bf0.57 & 0.29  & \bf 0.77  & \bf 0.71  &	 0.73  &  0.65 &  0.60  & 	\bf0.43  & \bf 0.52  & \bf 0.25  &	 0.59  &  0.71 \\ 

\name (32\textit{dim}) & \bf 0.62 &  \bf 0.85 & 	\bf0.53 & \bf0.24	 & \bf0.55 & \bf0.32  &  0.77  &  0.69  &	\bf 0.74  & \bf 0.66 & \bf 0.68  & 	0.41  & \bf 0.52  & \bf 0.25  &	\bf 0.61  & \bf 0.75 \\ \hline

\end{tabular}}
\caption{Macro F1 ($m\text{-}F1$) and Claim-F1 ($c\text{-}F1$) for ablation studies.}
\label{tab:ablation}
\end{table*}

The outputs of the POS and dependency embedding layers are subsequently fed to a BiLSTM and {\em GlobalAveragePooling} layers, respectively. Their respective outputs are projected to a 32-dimensional representation for the fusion.   
We employ HuggingFace's BERT implementation for computing the tweet representation. The 768-dimensional embeddings is projected to a 32-dimensional representation using linear layers. We progress with the 32-dimensional representation of BERT as we observe no elevation in results on using the 768-dimensional representation, as can be seen in Table \ref{tab:ablation}. Besides, the latter results in $\sim2$ million trainable parameters, whereas the former requires $\sim1.2$ million trainable parameters.
We employ sparse categorical cross-entropy loss with Adam optimizer and use softmax for the final classification.\footnote{ \newcite{gupta:lesa:claim:detection:eacl:2021:supplementary} accompanies other hyperparameters.} For evaluation, we adopt {\bf macro-F1 ($m\text{-}F1$)} and \textbf{claim-F1 ($c\text{-}F1$)} scores used by the existing methods \cite{daxenberger2017essence, chakrabarty2019imho}.



We perform our experiments in two setups. In the {\bf first in-domain setup}, we train, validate and test on the same dataset and repeat it for all seven datasets independently. In the {\bf second general-domain setup}, we combine all datasets and train a unified generic model. Subsequently, we evaluate the trained model on all seven datasets individually.

Furthermore, for each experiment, we ensure a balanced training set by down-sampling the dominant class at $1:1$ ratio. However, we use the original test set for a fair comparison against the existing baselines and state-of-the-art models.



\section{Experimental Results}
Table \ref{tab:ablation} shows $m\text{-}F1$ and $c\text{-}F1$ for different variants of \name. We begin with a fine-tuned BERT model and observe the performance on test sets of all seven datasets. On the Twitter dataset, the BERT architecture yields $m\text{-}F1$ score of 0.60 and $c\text{-}F1$ score of 0.83. We also report the weighted-average score as 0.58 $m\text{-}F1$ and 0.73  $c\text{-}F1$, in the last two columns of Table \ref{tab:ablation}. Since we hypothesize that claim detection has a strong association with the structure of the text, we amalgamate POS and dependency (DEP) information with the BERT architecture in a step-wise manner. The BERT+POS model reports an increase of 1\% $m\text{-}F1$ and $c\text{-}F1$ scores on the Twitter dataset. We observe similar trends in other datasets and the overall weighted-average score as well. We also perform experiments on other permutations, and their results are listed in Table \ref{tab:ablation}. Finally, we combine both POS and DEP modules with the BERT architecture (\textit{aka.} \name). It obtains improved results for most of the cases, as shown in the last row of Table \ref{tab:ablation}. The best result on average stands at 0.61 $m\text{-}F1$ and 0.75 $c\text{-}F1$ for the proposed \name\ model. 
This serves as a testament to our hypothesis, validating our assumption that
combining syntactic and semantic representations leads to better detection of claims.

\begin{table*}[t]
\centering
\resizebox{0.9\textwidth}{!}{
\begin{tabular}{l||c|c|c|c|c|c|c|c|c|c|c|c|c|c||c|c}
   \multirow{3}{*}{\bf Models} & \multicolumn{2}{c|}{\bf Noisy} & \multicolumn{4}{c|}{\bf Semi-Noisy} & \multicolumn{8}{c||}{\bf Non-Noisy} & \multicolumn{2}{c}{\multirow{3}{*}{\bf Wt Avg}} \\ 
\cline{2-15}
 & \multicolumn{2}{c|}{Twitter} & \multicolumn{2}{c|}{OC} & \multicolumn{2}{c|}{WTP} & \multicolumn{2}{c|}{MT} & \multicolumn{2}{c|}{PE} & \multicolumn{2}{c|}{VG} & \multicolumn{2}{c||}{WD} & \multicolumn{2}{c}{} \\ \cline{2-15}
 & $m\text{-}F1$ & $c\text{-}F1$ & $m\text{-}F1$ & $c\text{-}F1$& $m\text{-}F1$ & $c\text{-}F1$& $m\text{-}F1$ & $c\text{-}F1$& $m\text{-}F1$ & $c\text{-}F1$& $m\text{-}F1$ & $c\text{-}F1$& $m\text{-}F1$ & $c\text{-}F1$   \\ \hline
 
\hline  
BERT &	0.50	& 0.67&	0.50&	0.24&	0.36	&0.27&	0.75&	0.69&	0.73 &	\textbf{0.67} &	0.61 &	0.48&	0.48&	0.23 &	0.52	& 0.62 \\ 
XLNet &	0.52 &	0.70	& 0.45 &	0.24 &	0.55 &	0.30	& 0.49 &	0.43 &	0.71 &	0.64 &	0.53 &	0.43 &	0.51 &	0.12 &	0.54 &0.64 \\
\hline
Accenture	& 0.48 &	0.15&	0.44&	0.16&	0.50&	0.23&	0.48&	0.28&	0.45&	0.11&	0.39&	0.27&	0.34&	0.11&	0.46&	0.15\\

Team Alex &	\textbf{0.70}&	0.88&	0.46&	0.23&	0.52&	0.21&	0.75&	0.64&	0.69&	0.64&	0.32&	0.38&	0.60&	0.34&	0.59&	0.76\\

Check Square &	0.12&	0.02	& 0.49	& 0.25 &	0.39 &	0.26&	0.57&	0.32&	0.47&	0.11&	0.32	& 0.37 &	\textbf{0.76} &	\textbf{0.56} & 0.35 & 0.07 \\	

CrossDomain & 0.67	&	0.84 	& \textbf{0.61*}	&	\textbf{0.26*}	&	\textbf{0.59*}	&	0.29*	&	0.79*	&	0.67*	&	\textbf{0.74*}	&	0.61*	&	0.66*	&	0.45*	&	0.63*	&	0.29*	& \textbf{0.65}		& 0.74\\

CrossDomain$^{\dagger}$ & 0.67	&	0.84 & 0.50 & 0.24 & 0.52 & 0.27 & 0.85 & 0.79 & 0.71 & 0.63 & 0.60 & 0.46 & 0.59 & 0.31 &  0.61 & 0.74 \\

\hline
\name & 0.67 & \textbf{0.89} &	0.51 &	\textbf{0.26} &	0.57 &	\textbf{0.33} &	\textbf{0.80} &	\textbf{0.71} &	 0.73 &	\textbf{0.67} &	\textbf{0.68} &	\textbf{0.52} &	0.61 &	0.35&	0.63&	 \textbf{0.79}\\ \hline
\end{tabular}}
\caption{Macro F1 ($m\text{-}F1$) and F1 for claims ($c\text{-}F1$) in the in-domain setup. For CrossDomian, the asterisk (*) indicates results taken from \newcite{daxenberger2017essence} and the dagger (${\dagger}$) represents the \textit{\underline{reproduced}} results.} 
\label{Table: individual results}
\end{table*}

\begin{table*}[t]

\centering
\resizebox{0.9\textwidth}{!}{
\begin{tabular}{l||c|c|c|c|c|c|c|c|c|c|c|c|c|c||c|c}
   \multirow{3}{*}{\bf Model} & \multicolumn{2}{c|}{\bf Noisy} & \multicolumn{4}{c|}{\bf Semi-Noisy} & \multicolumn{8}{c||}{\bf Non-Noisy} & \multicolumn{2}{c}{\multirow{3}{*}{\bf Wt Avg}} \\ 
\cline{2-15}
 & \multicolumn{2}{c|}{Twitter} & \multicolumn{2}{c|}{OC} & \multicolumn{2}{c|}{WTP} & \multicolumn{2}{c|}{MT} & \multicolumn{2}{c|}{PE} & \multicolumn{2}{c|}{VG} & \multicolumn{2}{c||}{WD} & \multicolumn{2}{c}{} \\ \cline{2-15}
 & $m\text{-}F1$ & $c\text{-}F1$ & $m\text{-}F1$ & $c\text{-}F1$& $m\text{-}F1$ & $c\text{-}F1$& $m\text{-}F1$ & $c\text{-}F1$& $m\text{-}F1$ & $c\text{-}F1$& $m\text{-}F1$ & $c\text{-}F1$& $m\text{-}F1$ & $c\text{-}F1$   \\ \hline
 
\hline
BERT & 0.60 &	0.83 &	0.52 &	0.24 &	0.53&	\bf 0.32&	0.70 &	0.63&	0.69&	0.64&	0.58&	0.43&	0.48&	0.22& 0.58 	& 0.73 \\ \hline

XLNet & 0.59 &	0.81 &	0.56 &	\bf 0.28 &	\bf 0.57 &	0.29 &	0.68 &	\bf 0.69 &	0.71 &	0.64 &	0.61 &	\bf 0.44 &	\bf 0.52 &	\bf 0.25 & 0.59 & 0.72\\

Accenture & 0.49 &	0.43 &	0.31 &	0.12 &	0.40 &	0.18 &	0.36 &	0.13 &	0.51 &	0.36 &	0.38 &	0.17 &	0.37 &	0.04 &	0.43 &	0.38\\

Team Alex & 0.54 &	0.75 &	0.54 &	0.25 &	0.54&	0.30 &	0.71 &	0.65 &	0.71&	0.63&	0.61&	0.43&	0.48&	0.19&	0.57 &	0.67 \\

Check Square &	0.58&	0.82&	0.51&	0.23&	0.48&	0.28&	0.56&	0.53&	0.68&	0.59&	0.56&	0.38&	0.47&	0.21&  0.56  & 0.72 \\

CrossDomain & \bf 0.65&	0.82&	\bf 0.57&	0.27&	0.53&	0.28&	0.71&	0.63&	0.66&	0.57&	0.61&	0.43&	\bf 0.52&	\bf 0.25  & 0.60 & 0.71\\

\hline
\name & 0.62 &  \bf 0.85 & 	0.53 & 0.24	 & 0.55 & \bf0.32  & \bf 0.77  & \bf 0.69  &	\bf 0.74  & \bf 0.66 & \bf 0.68  & 	0.41  & \bf 0.52  & \bf 0.25  &	\bf 0.61  & \bf 0.75 \\ \hline

\end{tabular}}

\caption{Macro F1 ($m\text{-}F1$) and Claim-F1 ($c\text{-}F1$) in the general-domain setup.} 
\label{Table: combined result}
\end{table*}

In all aforementioned experiments, we use our pre-defined concept of three viewpoints, i.e., noisy, semi-noisy and non-noisy. Therefore, for completeness, we also construct a combined viewpoint which does not contain any structure-specific pillar in POS or DEP branches. The results from this ablation experiment are reported in \name\ (Combined-view) row. We observe that the combined-view results are inferior to the variant with separate viewpoints for each component 
(c.f. second last and last row of Table \ref{tab:ablation} respectively).
Thus, providing attention to datasets based on the noise in their content is demonstrated by a significant increase of $\sim2\%$ $m\text{-}F1$ from combined viewpoint to separate viewpoints experiment.

\subsection{Baselines and Comparative Analysis}
We employ the following baselines (some of them being state-of-the-art systems for claim detection and text classification):
$\triangleright$ \textbf{XLNet} \cite{yang2019xlnet}: It is similar to the BERT model, where we fine-tune XLNet for the claim detection;
$\triangleright$ \textbf{Accenture} \cite{williams2020accenture}: A RoBERTa-based system that ranked first in the CLEF-2020 claim detection task \cite{checkthat:clef:2020};
$\triangleright$ \textbf{Team Alex} \cite{nikolov2020team}: The second-ranked system at CLEF-2020 task that fused tweet meta-data into RoBERTa for the final prediction;
$\triangleright$ \textbf{CheckSquare} \cite{cheema2020checksquare}: An SVM-based system designed on top of pre-trained BERT embeddings in addition to incorporating POS and dependency tags as external features.
$\triangleright$ \textbf{CrossDomain} \cite{daxenberger2017essence}: Among several variations reported in the paper, their best model incorporates CNN (random initialization) for the detection.
We reproduce the top submissions from CLEF-2020 challenge using the best performing models mentioned in the referenced papers. Code for CheckSquare was provided online. For Accenture and Team Alex we reproduce their methods using the hyper-parameters mentioned in the paper. We evaluate all baselines using the same train and test set as for \name\ . 
\begin{table}[t]
\centering
\resizebox{0.4\textwidth}{!}
{
\begin{tabular}{l||c|c|c|c|c|c}
   \multirow{2}{*}{\bf Models} & \multicolumn{2}{c|}{\bf Noisy} & \multicolumn{2}{c|}{\bf Semi-Noisy} & \multicolumn{2}{c}{\bf Non-Noisy} \\ 
\cline{2-7}
 & $m\text{-}F1$ & $c\text{-}F1$ & $m\text{-}F1$ & $c\text{-}F1$ & $m\text{-}F1$ & $c\text{-}F1$ \\ \hline
 
\hline
BERT  & 0.60 & 	0.83  & 0.52 &	\bf 0.29 &	0.63 &	0.58 \\ \hline

XLNet & 0.59 &	0.81 & \bf 0.57 & \bf 0.29	& 0.65 & 0.59 \\

Accenture & 0.49 &	0.43 & 0.36 & 0.16 & 0.45 & 0.30 \\

Team Alex & 0.54 &	0.75 & 0.54 & 0.28 & 0.65 & 0.57 \\

CheckSquare & 0.58 &	0.82 & 0.49 & 0.26 &0.61 &0.53  \\

CrossDomain & \bf 0.65 & 0.82 & 0.55 & 0.28 & 0.63 & 0.53 \\
\hline
\name & 0.62 & \bf 0.85 & 0.54 & \bf 0.29 & \bf 0.69 &	\bf 0.60 \\\hline
\end{tabular}}
\caption{Category-wise weighted-average F1 scores.} 
\label{tab:category:avg}
\end{table}

\begin{table*}[ht!]
\centering
\resizebox{0.9\textwidth}{!}{
\begin{tabular}{l|l|p{30em}|c|c|c}
 \multicolumn{2}{c|}{} & \multirow{2}{*}{\bf Example}  & \multirow{2}{*}{\bf Gold}  & \multicolumn{2}{c}{\bf Prediction} \\ \cline{5-6}
 \multicolumn{2}{c|}{} & & & \bf \name & \bf CrossDomain \\ \hline

\hline
    \multirow{3}{*}{TWR} & $x_1$ & \textit{28 coronaoutbreak cases thus far in india italian tourists 16 their driver 1 kerala 3 cureddischarged agra 6 delhi 1 noida school dad telangana 1 coronavirusindia} & 1 & \textcolor{red}{0} & \textcolor{red}{0}    \\ \cline{2-6}
    & $x_2$ & \textit{can we just call this a cure now} & 0 & 0 & \textcolor{red}{1} \\ \hline
    
    \multirow{2}{*}{MT} & $x_3$ & \textit{Besides it should be in the interest of the health insurers to recognize alternative medicine as treatment, since there is a chance of recovery.} & 0  & \textcolor{red}{1} &  \textcolor{red}{1} 
    \\ \hline
     
    \multirow{2}{*}{PE} & $x_4$ & \textit{On the other hand, fossil fuels are abundant and inexpensive in many areas} & 0 & \textcolor{red}{1} & \textcolor{red}{1} \\ \cline{2-6}
     & $x_5$ & \textit{Daily exercise will help also to develop children's brain function.} & 1 & 1 & \textcolor{red}{0}  
    
    \\ \hline
    
    \multirow{3}{*}{OC} & $x_6$ & \textit{Skinny Puppy is headlining Festival Kinetik !} & 0 & \textcolor{red}{1} & \textcolor{red}{1} \\ \cline{2-6}
    & $x_7$ & \textit{I guess I'm not desensitized enough to just forget 
    about people being murdered in my neighborhood.} & 1 & 1 & \textcolor{red}{0} \\ \hline    
    WD & $x_{8}$ & \textit{No wonder 50 million babies have been aborted since 1973 .} & 0 & \textcolor{red}{1} &  \textcolor{red}{1}\\ \hline
\end{tabular}}

\caption{Error analysis of the outputs. Red texts highlight errors.}
\label{tab:error:examples}
\end{table*}

We report our comparative analysis for the in-domain setup in Table \ref{Table: individual results}.
We observe that \name\ obtains best $c\text{-}F1$ scores for six out of seven datasets. Additionally, it achieves a weighted average $c\text{-}F1$ of $0.79$ which is $3.95\%$ improvement over the best performing baseline. In terms of $m\text{-}F1$ values, our weighted average ranks second next to CrossDomain. We reproduced CrossDomain baseline using their GitHub code \cite{ukplab}. If the reproduced values are considered, our model outperforms all other models in $m\text{-}F1$ value as well.

Similarly, we compile the results for the general-domain setup in Table \ref{Table: combined result}.
In the non-noisy category, \name\ obtains better $m\text{-}F1$ scores than three of the four state-of-the-art systems, i.e., it reports $0.77$, $0.74$, and $0.68$ $m\text{-}F1$ scores compared to $0.71$, $0.71$, and $0.61$ $m\text{-}F1$ scores of the comparative systems on MT, PE, and VG test sets, respectively. On WD, we observe similar $m\text{-}F1$ and $c\text{-}F1$ scores for both the best baseline and \name. On the datasets in other categories, we observe comparative $m\text{-}F1$ scores; however, none of the baselines are consistent across all dataset -- e.g., CrossDomain \cite{daxenberger2017essence} reports the best $m\text{-}F1$ scores on Twitter and OC, but yields (joint) fourth-best performance on WTP. Moreover, \name\ yields the best $m\text{-}F1$ score across the seven datasets on average with $\ge1\%$ improvements. On the other hand, we obtain best $c\text{-}F1$ scores for five out of seven datasets. In addition, \name\ reports overall $c\text{-}F1$ of 0.75 with a significant improvement of $\ge3\%$. 
Using a paired T-test, \name\ showed significant statistical improvement compared against BERT in $m\text{-}F1$ and $c\text{-}F1$ for the noisy dataset with p-values $.00017$ and \textless $.00001$ respectively. Results were also significant for $m\text{-}F1$ and $c\text{-}F1$ for PE and $m\text{-}F1$ for WD. The small sample size in some datasets like MT and VG does not allow a reliable calculation of test statistics.

Since our work intends to developing a model that is able to detect claims irrespective of the source and origin of text, we also analyse the weighted-average scores for each category in Table \ref{tab:category:avg}. We observe that \name\ obtains the best $c\text{-}F1$ scores in each category, in addition to the best $m\text{-}F1$ score in non-noisy category as well. For the other two categories, \name\ yields comparative performances. The results are better for noisy data than for non-noisy owing to the small size and skewness against claims in the latter’s test set. Therefore, misclassification of a single claim causes severe penalization to $c\text{-}F1$.

\subsection{Error Analysis}
It is apparent from the results that all systems (including \name) committed some errors in claim detection. Thus, in this section, we  explore where our system misclassified the inputs by analysing some examples. Table \ref{tab:error:examples} presents a few instances along with the gold labels and the predictions of the best-performing  baseline, CrossDomain \cite{daxenberger2017essence}, for comparison.  In some cases, both \name\ and CrossDomain failed to classify the instances correctly, whereas in others, \name\ classifies the instances correctly but CrossDomain could not. 
We also report intuitions for the misclassification by \name\ in some cases. The presence of numbers and statistics could be the reason behind the misclassifications in examples $x_1$ and $x_{8}$. Example $x_3$ contains two weak phrases (\textit{`alternative medicine as treatment'} and \textit{`there is a chance of recovery'}) which are most likely the cause of misclassification. The former might have been interpreted as suggestion backed up by some evidence, while in the latter phrase, \name\ might have misinterpreted the optimism with claim. Furthermore, the phrase \textit{`fossil fuels are abundant'} in example $x_4$ reflects world knowledge instead of a claim, as interpreted by \name.

\section{Conclusion}
In this paper, we addressed the task of claim detection from online posts. To do this, we proposed a generic and novel deep neural framework, \name, that leverages the pre-trained language model and two linguistic features, corresponding to the syntactic properties of input texts, for the final classification. Additionally, we tackled the texts from distinct sources for the claim detection task in a novel way. In particular, we categorized the input text as noisy, non-noisy, and semi-noisy based on the source, and modeled them separately. Subsequently, we fused them together through an attention module as the combined representation.     

One of the major bottlenecks of claim detection in online social media platforms is the lack of qualitative annotation guidelines and a sufficiently large annotated dataset. Therefore, we developed a large Twitter dataset of $\sim10,000$ manually annotated tweets for claim detection. In addition to our twitter dataset, we employed six benchmark datasets (representing either semi-noisy or non-noisy input channels) for evaluation of the proposed model. We compared the performance of \name\ against four state-of-the-art systems and two pre-trained language models. Comparison showed the superiority of the proposed model with $\ge3\%$ claim-F1 and $\ge1\%$ macro-F1 improvements compared to the best performing baselines on average. As a by-product of the study, we released a comprehensive guideline for claim annotation.

\section*{Acknowledgement}
We would like to thank Rituparna  and  LCS2 members for helping in data annotation. The work was partially supported by Accenture Research Grant,   Ramanujan Fellowship, and CAI, IIIT-Delhi. 

\bibliography{eacl2021}
\bibliographystyle{acl_natbib}

\end{document}


\maketitle

\appendix


\appendix
\section{Datasets}
\label{sec:dataset}

There exist a few datasets \cite{peldszus-stede-2015-joint,stab-gurevych-2017-parsing} for claim detection in online text; however, most of them are formal and structured texts. Despite the abundance of tweets, literature does not suggest any significant effort for claim detection in Twitter, and arguably, the prime reason is the lack of large-scale dataset. 

Therefore, we attempted to develop a new and relatively larger dataset for claim detection in OSM platforms. We collected $\sim 40,000$ tweets from various sources \cite{carlson_2020,smith_2020,celin_2020,
info:doi/10.2196/19273,QCRI} and manually annotated them. We additionally included claim tweets of \newcite{alam2020fighting} and CLEF-2020 \cite{checkthat:clef:2020}. 

To the best of our knowledge there exists only one relevant study (by \cite{alam2020fighting}) that proposed guidelines for annotating claims in tweets. Some problems with the guidelines are as follows:
\begin{itemize}
    \item The authors classified tweets as factually verifiable claims and non-factually verifiable claims, which is only a subset of the claims that exist.
    \item They do not consider personal opinions as claims. Therefore the tweet \textit{``im actually starting to feel like europe is trying to make sure the virus spreads in africa"} has been labelled non-claim. This is problematic because personal opinions with societal implications have the potential to result in conspiracy theories and cause public unrest.
    \item They label a tweet as a claim only if the entire sentence is a claim; claims existing in a sub-sentence or sub-clause is not considered as a claim. However from this example \textit{``note to with covid19 you cant declare bankruptcy settle out of court pay it to keep quiet hope it disappears in the spring ignore deaths claim its a hoax covid19"}, which has been labelled non-claim, it is clear that claims can be made in sub-clauses (the claim here being covid-19 is a hoax) and are equally important to detect and verify.
    \item They do not consider indirect claims. For example the following tweet is labelled non-claim: \textit{``folks when you say the corona virus isnt a big deal it only kills the disabled elderly chornicallyill and immunocompromised the implication is that those people are expendable please be more careful"} 
    This tweet indirectly implies that corona only kills the disabled, elderly, chronically ill and immunocompromised persons which is clearly a claim. 
    \item They do not consider claims made in sarcasm or humour. For example, \textit{``crap this virus is turning all the people into pigeons coronavirus"} is labelled as a non-claim under their guidelines.
\end{itemize}
  
The aforementioned reasons motivated us to form our own set of guidelines that are more exclusive, nuanced and applicable to our understanding of a claim.

In this section, we present how we annotated the tweets, extrapolating the set of extensive guidelines we formed for the process and the pre-processing methods we adopted.

\subsection{Data preprocessing}
Before annotating the dataset, we perform the preliminary task of data cleaning. Our pre-processing stage involves removing hashtags, URLs, user handles and non-ASCII characters. All tweets with character count less than $20$ and word count less than $4$ are also removed, owing to the lack of context for their interpretation. Finally, we spell-check the words using symspellpy\footnote{https://pypi.org/project/symspellpy/}. The final dataset contains $9,894$ tweets which were split into $70:15:15$ for training, validation and testing.

\subsection{Data Annotation Guidelines}

Our official definition adopted for claims is to {\em state or assert that something is the case, with or without providing evidence or proof}. Following are our guidelines for what qualifies as a claim. Anything that does not qualify as a claim, is labelled as `non-claim'. However, certain clarifying guidelines are also given for non-claims in Section \ref{guidelines: non-claims}. 

Note that some guidelines for claims also contain situations (with examples) where the guidelines do not generalise and input is labelled non-claim; vice versa also holds. Such exceptions are preceded by an asterisk (*).

\textit{Labels}: The tweets are labelled as 1 for claim, 0 for non-claim and x in obscure situations. This is considered only when the language of the tweet is incomprehensible or if no guideline can be referred to annotate it.

\subsection{Guidelines for `claims'}\label{guidelines: claims}

\begin{itemize}
    \item Tweets mentioning \underline{statistics}, dates or numbers. \\
\textbf{Example:}    
\textit{[``just 1 case of corona virus in india and people are crazy for masks daily 400 people die in road crashes still no craze for helmetsthinking face safetysaves be it virus or road crashes"]}

    \item Tweets mentioning a \underline{personal experience}. \\
    \textbf{Example:} 
\textit{[``i live in seattle i have all symptoms of covid19 and have a history of chronic bronchitis since i work in a physical therapy clinic with many 65 patients and those with chronic illnesses i decided to be responsible and go to get tested this is how that went"]}

    \item Tweets \underline{‘reporting’ something to be true} or an instance to have happened or will happen. \\
    \textbf{Example:} 
\textit{[``breaking boris johnson says he visited kettering hospital shook hands with corona patients but the hospital doesnt have corona cases shaking hands would be dangerous not sure how ill gloss over the fact the pm is a liar a complete fucking idiot but ill find a way x"]}

    \item Tweets \underline{containing verified facts} also account for a claim, a veracious claim that is. \\
    (\textit{Note}: a fact known by one, may not necessarily be known by another)\\
    \textbf{Example:} 
\textit{[``The Chinese CDC has started research and development of a vaccine for the \#coronavirus.”]} - known fact

    \item Tweets that \underline{negate a claim} are also accounted as claims.\\
    \textbf{Example:} 
\textit{[``disinfectants are not a cure for coronavirus”]}

    \item Tweets that \underline{indirectly (subtly) imply} that something is true. \\
    \textbf{Example 1:} 
\textit{[``b52questions 1 why is rudy not under arrest 2 why is harvey not in rikers 3 why is cuccinelli still working 4 has barr quit yet 5 when is flynn being sentenced 6 who trusts pence and mrs miller with messaging about coronavirus 7 are rs happy wtheir guy"]} - indirectly implies rudy is not under arrest and harvey is not in rikers\\
\textbf{Example 2:} 
\textit{[``do rich people know theyll get the virus if poor people cant be tested and diagnoseddo rich people know theyll get the virus if poor people cant be tested and diagnosed"]} - indirectly implies rich people will get the virus if poor people can’t be tested

    \item Claims \underline{made in sarcasm or humour}.\\
    \textbf{Example 1:} 
\textit{[``@TheDailyShow Newsflash! \#trumpfact If you paint your face \#orange you will be \#immune to \#coronavirus”]} \\
\textbf{Example 2:} 
\textit{[``RT @\_saraellen: If you’ve ever used messers bathroom you’re immune to corona virus.”]}\\
\textbf{Example 3:} 
\textit{[``corona virus minding its business by avoiding africa and going to other continents"]} \\
Example 1 and 2 are both examples of claims, even if evidently sarcastic; Example 3 is a humour oriented claim.

\item All opinions are not claims. \underline{Opinions that have societal implications} are considered as claims.\\
\textbf{Example 1:} 
\textit{[``@derekgilbert I think the Chinese stole a bio weapon https://t.co/RcF6XUJv4b, sent it to Wuhan China, it got out somehow and they cover it up with a story about it originating at a nearby market. They know how bad the virus is and quarantine entire cities. https://t.co/ZwMqsiWGau"]} \\

\textbf{Example 2:} 
\textit{[``I think Burger King fries are better than Mc'D's"]}\\ 
Example 1 is an opinion that claims something to have happened, whose veracity will affect a certain section of the society. Hence it will be marked as a claim. Example 2, on the other hand, is a personal belief that majorly impacts only the person making the tweet and is hence marked non-claim.  

\item Tweet that says something is true and provides an \underline{attachment as evidence} or to support the statement. \\
\textbf{Example 1:} 
\textit{[``RT @Jawn42: If you ate here growing up, you're immune to the Coronavirus. https://t.co/b9a0hm171b"]}\\
\textbf{Example 2:} 
\textit{[``RT @TerminalLance: This kills Coronavirus in the system https://t.co/iOFNlkSrUj"]}\\  \\
\textbf{{*}} However, if a person says something is provided in the attachment, that will not be a claim. \\
\textbf{Example:}
\textit{[``im stunned by the depth of coronavirus information being released in singapore on this website you can see every known infection case where the person lives and works which hospital they got admitted to and the network topology of carriers all laid out on a timeseries link"]}

\item A claim can be a sub-part of a question. \\
\textbf{Example:} 
\textit{[``Does the pneumonia shot help protect from developing pneumonia caused by \#covid19"]} - the claim here being that pneumonia is caused by COVID-19.
\end{itemize}

\begin{table*}[t]

\centering
\resizebox{\textwidth}{!}{
\begin{tabular}{l||c|c|c|c|c|c|c|c|c|c|c|c|c|c||c|c}
   \multirow{3}{*}{\bf Models} & \multicolumn{2}{c|}{\bf Noisy} & \multicolumn{4}{c|}{\bf Semi-Noisy} & \multicolumn{8}{c||}{\bf Non-Noisy} & \multicolumn{2}{c}{\multirow{3}{*}{\bf Wt Avg}} \\ 
\cline{2-15}
 & \multicolumn{2}{c|}{Twitter} & \multicolumn{2}{c|}{OC} & \multicolumn{2}{c|}{WTP} & \multicolumn{2}{c|}{MT} & \multicolumn{2}{c|}{PE} & \multicolumn{2}{c|}{VG} & \multicolumn{2}{c||}{WD} & \multicolumn{2}{c}{} \\ \cline{2-15}
 & $m\text{-}F1$ & $c\text{-}F1$ & $m\text{-}F1$ & $c\text{-}F1$& $m\text{-}F1$ & $c\text{-}F1$& $m\text{-}F1$ & $c\text{-}F1$& $m\text{-}F1$ & $c\text{-}F1$& $m\text{-}F1$ & $c\text{-}F1$& $m\text{-}F1$ & $c\text{-}F1$   \\ \hline
 
\hline
POS-only [2-gram] &	0.41 &	0.59 &	\bf0.52 &	0.14 &	\bf0.54 &	0.23 &	0.42 & 0.00 & 0.43 & 0.06 &	\bf0.55 &	0.30 &	\bf0.50 & 0.09 & 0.47 & 0.47\\

POS-only [3-gram] & \bf 0.49 &	\bf 0.74 &	0.51 &	\bf0.17 &	0.49 &	\bf0.24 &	0.54 &	0.32 &	\bf0.51 &	\bf0.32 &	\bf0.55 &	\bf0.39 &	0.43 &	\bf0.10 & \bf0.50 & \bf0.62\\


POS-only [4-gram] & 0.23 &	0.23 &	0.50 &	0.05 &	0.53 &	0.14 &	\bf 0.61 &	\bf0.35 &	0.43 &	0.06 &	0.53 &	0.18 &	0.47 &	0.00 & 0.41 & 0.19 \\
\hline

\end{tabular}}
\caption{Macro F1 and claim-F1 for POS n-gram experiments. }
\label{Table: POS n-gram}
\end{table*}
\begin{table*}[t]

\centering
\resizebox{\textwidth}{!}{
\begin{tabular}{l||c|c|c|c|c|c|c|c|c|c|c|c|c|c||c|c}
   \multirow{3}{*}{\bf Models} & \multicolumn{2}{c|}{\bf Noisy} & \multicolumn{4}{c|}{\bf Semi-Noisy} & \multicolumn{8}{c||}{\bf Non-Noisy} & \multicolumn{2}{c}{\multirow{3}{*}{\bf Wt Avg}} \\ 
\cline{2-15}
 & \multicolumn{2}{c|}{Twitter} & \multicolumn{2}{c|}{OC} & \multicolumn{2}{c|}{WTP} & \multicolumn{2}{c|}{MT} & \multicolumn{2}{c|}{PE} & \multicolumn{2}{c|}{VG} & \multicolumn{2}{c||}{WD} & \multicolumn{2}{c}{} \\ \cline{2-15}
 & $m\text{-}F1$ & $c\text{-}F1$ & $m\text{-}F1$ & $c\text{-}F1$& $m\text{-}F1$ & $c\text{-}F1$& $m\text{-}F1$ & $c\text{-}F1$& $m\text{-}F1$ & $c\text{-}F1$& $m\text{-}F1$ & $c\text{-}F1$& $m\text{-}F1$ & $c\text{-}F1$   \\ \hline
 
\hline
POS-only [Bi-LSTM] & \bf 0.49 &	\bf 0.74 &	\bf 0.51 &	\bf0.17 &	\bf0.49 &	\bf0.24 &	0.54 &	0.32 &	0.51 &	0.32 &	0.55 &	\bf0.39 &	\bf0.43 &	0.10 & \bf0.50 & \bf0.62\\

POS-only [transformer] & 0.48 & 0.72 &	0.43 &	\bf 0.18 &	0.42 &	0.20 &	\bf0.56 &	\bf0.37 &	\bf0.57 &	\bf0.36 &	\bf0.58 &	0.35 &	\bf0.43 &	\bf0.20 & 0.48 & 0.61\\
\hline

\end{tabular}}
\caption{Macro F1 and claim-F1 for POS architecture experiments}
\label{tab:pos:arch:variants}
\end{table*}

\subsection{Guidelines for `non-claims'} \label{guidelines: non-claims}

\begin{itemize}
    \item Hoping that something happens or feeling something is true is not claiming it.\\
    \textbf{Example 1:} 
\textit{[``World doesn’t end if u don’t give your opinions about corona virus. I’m drinking  \#nilavembu boiled in hot water and hope it prevents. \#COVID-19 \#coronapocolypse”]}\\
\textbf{Example 2:} 
\textit{[``I feel like I’m immune to coronavirus.”]}

    \item Inclusion of words that project doubt over the said statement.\\
    \textbf{Example:} 
\textit{[``\@politicalelle Political correctness infecting the \#Coronavirus . Let's change words describing the virus maybe that will cure it.”]}\\\\
\textbf{{*}} Tweets containing doubt-casting words can still, however, contain claims.\\
\textbf{Example:} 
\textit{[``Coronavirus may have originated in lab linked to China's biowarfare program \#coronavirus https://t.co/2NSWidMkoa”]}

    \item Urging one to not claim something or to spread misinformation is not a claim.\\
    \textbf{Example:} 
\textit{[``\#Covid19 - Dear all: Stop telling the public that plaquenil/Azithromycine is a cure!!!!! Plz some leadership is needed regarding this matter!”]}

    \item Questioning a possible claim is not a claim.\\
    \textbf{Example:} 
\textit{[``Do disinfectants really cure Corona?”]}\\\\
\textbf{{*}} However, a tweet containing a question can still comprise a claim.\\
\textbf{Example:} 
\textit{[``Would you like to promote a cure that can kill 90\% \#COVID-19 virus in the body in 3-15 min?"]}\\
The above tweet claims that there exists a cure that can kill 90\% of COVID-19 virus in the body

    \item Warning someone against a claim is not a claim.\\
    \textbf{Example:} 
\textit{[``If you think drinking disinfectants will cure \#Covid\_19 , you deserve death \#trump”]}\\
The above tweet may be hate speech but it does not say something is true or false. Hence it does not fall under the jurisdiction of claims.

\end{itemize}

\section{Experimental Details}
In this section, we report a few additional experimental results and other supplementary details.

\subsection{POS embeddings - Trigram}
We compute POS embeddings by learning word2vec skip-gram model \cite{mikolov2013efficient} on the tri-gram POS sequence. The choice of tri-gram sequence is empirical. Table \ref{Table: POS n-gram} shows results of our comparative experiments between bi-gram, tri-gram and four-gram sequences. As can be observed, $c\text{-}F1$ score for tri-grams is highest for six out of seven datasets with a weighted average increase of $\ge30\%$ over bi-grams and a three-fold increase over four-grams. 
Weighted Average $m\text{-}F1$ for tri-grams is also the highest by a margin of $6.38\%$ over the next best performing bi-grams.

\subsection{POS embeddings - Bi-LSTM}
In Table \ref{tab:pos:arch:variants}, we report the experimental results to support the choice of using Bi-LSTM, as oppose to Transformers, for extracting the POS features. As evident from Table \ref{tab:pos:arch:variants}, both $c\text{-}F1$ and $m\text{-}F1$ scores for Bi-LSTM are better for 4 out of 7 datasets. Additionally, the weighted average for both metrics is higher in case of Bi-LSTM.

\subsection{Hyperparameter}
Detail about hyperparameters is given in Table \ref{tab:hyperparams}. For the skip-gram model, we set \textit{context window} $=6$, \textit{embedding dimension} $=20$, and discard the POS sequence with \textit{frequency} $\le2$. 

The DEP Embedding for each text distribution is prepared using the transformer architecture, wherein for our specific prototype, we compute dependency embeddings with \textit{dimension} $=20$ using $5$ attention heads and a feed-forward \textit{dimension} $=128$. The attained representation is then pooled using {\em GlobalAveragePooling} and then passed through two linear layers with $64$ and $32$ hidden units respectively.


The BERT Embedding is trained as a downstream task wherein we use "bert-base-uncased" provided as pre-trained Language Model by HuggingFace. We use an Adam Optimizer with a \textit{learning rate} $=2e^{-5}$ and a \textit{batch size} $=16$ for the same. The BERT-layer from after being fine-tuned on our corpus is kept frozen in the final model, while the pooled output from the same is passed through two dense layers with $768$ (default hidden size of the BERT configuration made available through Hugging Face) and $32$ hidden units respectively to obtain a representation for further use in the model.


\begin{table}[t!]
    \centering
    \begin{tabular}{p{9em}|c}
        \bf Hyper-parameter & \bf Config \\ \hline
        
        \hline
        
         \multicolumn{2}{c}{BERT Fine-tuning} \\ \hline
         Epochs& 3 \\ 
        Optimizer& Adam \\ 
         Learning Rate & $2e^{-5}$ \\ 
         Batch size & 16 \\ \hline
         \multicolumn{2}{c}{Dependency Encoder - Pretraining} \\ \hline
         Epochs & 10\\
         Embedding dimension & 20\\ 
         Attention heads & 5\\
         Feed-Forward units & 128 \\
         Dropout & 0.3 \\
         Optimizer & Adam\\
         Activation & ReLU \& SoftMax\\
         Loss function & Cross Entropy\\\hline
         \multicolumn{2}{c}{Part-of-Speech - Pretraining} \\ \hline
         Embedding dimension & 20\\
         Window span & 6\\
         Minimum count & 2\\ \hline
         \multicolumn{2}{c}{Amalgamated Model} \\ \hline
         Epochs & 25\\
         Batch size & 256\\
         Bi-LSTM units & 128\\
         Hidden units & 256, 32, 16, 8\\
         Dropout & 0.3\\
         Optimizer & Adam\\
         Activation & ReLU \& SoftMax\\
         Loss function & Cross Entropy\\\hline
         
         \hline
    \end{tabular}
    \caption{Hyper-parameters of our \name\ model.}
    \label{tab:hyperparams}
\end{table}

The information state obtained from the concatenation of the prior three representations is then processed by an
attention layer followed by a dense layer of 16 units and ReLU activation function and $30\%$ dropout for regularisation. Finally, two dense layers of $8$ units and $2$ units respectively culminate into a softmax for classification. The layers use ReLU and Softmax activation functions respectively. We use the Adam Optimizer and sparse categorical cross-entropy as the loss function for the main output as well as for the auxiliary outputs.

\bibliography{reference}
\bibliographystyle{acl_natbib}